\documentclass{article}

\usepackage{microtype}
\usepackage{graphicx}
\usepackage{subfigure}
\usepackage{booktabs} 

\usepackage{bbm}
\usepackage{amssymb}
\usepackage{amsmath}  
\usepackage{amsthm}
\DeclareMathOperator{\E}{\mathbb{E}}

\theoremstyle{definition}
\newtheorem{proposition}{Proposition}[section]

\usepackage{hyperref}

\usepackage[accepted]{icml2020}

\icmltitlerunning{Estimating Risk and Uncertainty in Deep Reinforcement Learning}

\begin{document}

\twocolumn[
\icmltitle{Estimating Risk and Uncertainty in Deep Reinforcement Learning}



\icmlsetsymbol{equal}{*}

\begin{icmlauthorlist}
\icmlauthor{William R. Clements}{indust}
\icmlauthor{Bastien Van Delft}{indust}
\icmlauthor{Beno\^{i}t-Marie Robaglia}{indust}
\icmlauthor{Reda Bahi Slaoui}{indust,ponts}
\icmlauthor{S\'{e}bastien Toth}{indust}
\end{icmlauthorlist}

\icmlaffiliation{indust}{Indust.ai, Paris, France}
\icmlaffiliation{ponts}{Ecole des Ponts Paristech, Champs-sur-Marne, France}

\icmlcorrespondingauthor{William R. Clements}{mail@william-clements.com}

\icmlkeywords{Machine Learning, Reinforcement Learning, Uncertainty Estimation}

\vskip 0.3in
]



\printAffiliationsAndNotice{} 

\begin{abstract}
Reinforcement learning agents are faced with two types of uncertainty. Epistemic uncertainty stems from limited data and is useful for exploration, whereas aleatoric uncertainty arises from stochastic environments and must be accounted for in risk-sensitive applications. We highlight the challenges involved in simultaneously estimating both of them, and propose a framework for disentangling and estimating these uncertainties on learned Q-values. We derive unbiased estimators of these uncertainties and introduce an uncertainty-aware DQN algorithm, which we show exhibits safe learning behavior and outperforms other DQN variants on the MinAtar testbed.
\end{abstract}

Distinguishing between both \textit{epistemic} uncertainty, which stems from limited data, and \textit{aleatoric} uncertainty, caused by intrinsic stochasticity in the environment, is important in reinforcement learning for both exploration and risk-sensitivity  \cite{osband2016risk,moerland2017efficient,nikolov2019information}. However, while prior work has developed independent methods for estimating both uncertainties, difficulties appear when trying to estimate both simultaneously. For example, distributional reinforcement learning \cite{morimura2010nonparametric,morimura2012parametric,bellemare2017distributional}, which aims to learn the distribution of returns instead of the mean value only, has been suggested as a way of measuring aleatoric uncertainty \cite{nikolov2019information}. However, although metrics such as the variance of the learned distribution can be good indicators of the aleatoric uncertainty for in-distribution data, it was highlighted for example in \cite{chua2018deep} that for out-of-distribution data, when the epistemic uncertainty is high, such a metric is not a good indicator of aleatoric uncertainty as it conflates both uncertainties.

To address this issue, we construct a framework for disentangling both types of uncertainties which is applicable to both stochastic and deterministic environments. Our method builds on the distributional reinforcement learning framework, which aims to learn the entire return distribution instead of only its expected value \cite{bellemare2017distributional}, and methods for approximate Bayesian deep learning. Our main contributions are 1) a theoretical framework within which epistemic and aleatoric uncertainties can be separately estimated, 2) practical, unbiased estimators for both types of uncertainty, and 3) a demonstration that these uncertainties can successfully be used within an uncertainty-aware Deep Q Networks \cite{mnih2015human} algorithm.

\section{Background}

We consider a discounted Markov Decision Process (MDP) defined by $(\chi,A,R,P,\gamma)$, in which $\chi$ and $A$ represent the state and action spaces, $R$ is the distribution of rewards associated with performing actions given the states, $P$ is the transition probability, and $\gamma$ is the discount factor. 

Distributional reinforcement learning aims to learn the distribution of returns $Z^\pi(s,a)$ associated with taking action $a$ in state $s$ and then following a policy $\pi$. To learn this distribution, \cite{dabney2017distributional} propose a quantile parameterization. In this framework, a probability distribution $Z(s,a)$ is parameterized by $N$ quantiles $\tau_i=i/(N+1)$ for $i \in [1,N]$, with values $\boldsymbol{q} = (q_1,...,q_N)$. Learning the quantile values proceeds by minimizing the quantile regression loss \cite{koenker2001quantile},
\begin{align}
\label{quantileloss}
    &\mathcal{L}_q(\boldsymbol{q}) = \E_{z \sim Z(s,a) } \sum_{i=1}^N \rho_{\tau_i}(z - q_i(s,a)), \\
    &\text{ where}  \quad \rho_{\tau_i}(u) = u \times (\tau_i - \mathbbm{1}_{u<0}) \nonumber
\end{align} 
This loss can be minimized stochastically for each new value $z$ sampled from $Z(s,a)$. For temporal difference learning of the optimal value function, $Z(s,a)$ is replaced with the Bellman target $R(s,a) + \gamma Z(s',a')$, where $a' \sim \text{argmax}_{a'} \E [Z(s',a')]$ and $s' \sim P(\cdot |s,a)$, yielding the QR-DQN algorithm of \cite{dabney2017distributional}.

An intuitive way of estimating the aleatoric uncertainty on the return distribution would be to use the variance of the quantiles. However, the variance of the quantiles is not a good estimator of the aleatoric uncertainty, because for out of distribution data the epistemic uncertainty on the value of the quantiles can also affect the variance.

\section{Estimating both uncertainties}

Here, we construct a theoretical framework which will allow us to disentangle epistemic and aleatoric uncertainties and derived unbiased estimators for both.

\subsection{Theoretical framework}

We start by framing learning the quantiles of the return distribution as a Bayesian inference problem. We consider state $s$, action $a$ taken in state $s$, policy $\pi$, and data $D$ consisting of $K$ samples $(z_1,...,z_K)$ from $Z^\pi(s,a)$. To learn the value of a given quantile $\tau$ of $Z^\pi(s,a)$, we consider a neural network with parameters $\boldsymbol{\theta}$, which returns a value $y(\boldsymbol{\theta},s,a)$. We interpret possible values of $\boldsymbol{\theta}$ as different hypotheses about the function relating the state-action pair to the value of quantile $\tau$ of $Z^\pi(s,a)$ \cite{mackay2003information}. Following \cite{yu2001bayesian}, we define a likelihood based on how well the output of the network matches the data using an asymmetric Laplace distribution,
\begin{align}
P(D|\boldsymbol{\theta}) = \prod_{j=1}^K f_\tau(z_j - y(\boldsymbol{\theta},s,a)), \\ \text{where} \quad f_\tau(u) = \frac{\tau(1-\tau)}{\sigma_D} \exp(- \frac{\rho_\tau(u)}{\sigma_D}) \nonumber
\end{align}
\noindent where $\sigma_D$ is a characteristic length scale and $\rho_\tau$ is the same as in equation \ref{quantileloss}. 

To estimate the entire return distribution instead of a single quantile, we extend this formalism to a network with $N$ outputs $y_i(\boldsymbol{\theta},s,a)$, where each output $i$ is trained to learn the value of quantile $\tau_i$. We thus define the likelihood
\begin{align}
& P(D|\boldsymbol{\theta}) = \prod_{j=1}^K \prod_{i=1}^N f_{\tau_i}(z_j - y_i(\boldsymbol{\theta},s,a)) 
\label{likelihood}
\end{align}
Minimizing the loss in equation \ref{quantileloss} is equivalent to maximizing the likelihood in equation \ref{likelihood}. If we now consider a normal prior on parameters $\boldsymbol{\theta}$ centered around $0$, we can use any one of several methods for approximately sampling from the posterior distribution $P(\boldsymbol{\theta}|D)$ \cite{blundell2015weight,gal2016dropout,pearce2018bayesian}.

\subsection{Uncertainty estimates}

Using the framework described above, we now propose expressions for both aleatoric and epistemic uncertainties.

\subsubsection{Epistemic uncertainty}

To obtain a single aggregate measure of the epistemic uncertainty on the return distribution, we propose taking the average of the epistemic uncertainty on the quantiles, defined by their variance over $\boldsymbol{\theta}$,
\begin{align}
    \sigma_{\text{epistemic}}^2 = \E_{i \sim \mathcal{U}\{1,N\}} \left[ \text{var}_{\boldsymbol{\theta} \sim P(\boldsymbol{\theta}|D)} (y_i(\boldsymbol{\theta},s,a)) \right]
\label{epistemicdef}
\end{align}
where $\mathcal{U}\{1,N\}$ is the uniform distribution over $\{1,N\}$.

\subsubsection{Aleatoric uncertainty}

An intuitive measure of the aleatoric uncertainty is the variance of the quantile values. However, this variance is also affected by epistemic uncertainty in the form of the distribution over $\boldsymbol{\theta}$. To decouple aleatoric uncertainty from epistemic uncertainty, we define the aleatoric uncertainty as the variance of the expected value of the quantiles according to the posterior distribution over $\boldsymbol{\theta}$,
\begin{align}
    \sigma_{\text{aleatoric}}^2 = \text{var}_{i\sim \mathcal{U}\{1,N\}}[\E_{\boldsymbol{\theta}\sim P(\boldsymbol{\theta}|D)}y_i(\boldsymbol{\theta},s,a)]
\label{aleatoricdef}
\end{align}
When the posterior is concentrated around a single value, we recover the intuitive definition of aleatoric uncertainty as the variance of the quantiles. However, when the posterior is not concentrated, the variance of a single set of quantiles is a biased estimator of $\sigma_{\text{aleatoric}}^2$:
\theoremstyle{definition}
\begin{proposition}{}\textit{(Proof in the appendix)} Consider $\hat{\boldsymbol{\theta}}$ drawn from the posterior distribution over $\boldsymbol{\theta}$. Then $\text{var}_{i\sim \mathcal{U}\{1,N\}}[y_i(\hat{\boldsymbol{\theta}},s,a)]$ is a biased estimator of $\sigma_{\text{aleatoric}}^2$.
\end{proposition}

\subsubsection{Decomposition of Uncertainties}

We require that the total uncertainty on the return distribution can be decomposed as the sum of these two uncertainties. We consider the total variance of the return distribution $\text{var}_{\boldsymbol{\theta} \sim P(\boldsymbol{\theta}|D),i \sim \mathcal{U}\{1,N\} }(y_i(\boldsymbol{\theta},s,a))$, which for notational simplicity we write $\text{var}_{\boldsymbol{\theta},i}(y_i(\boldsymbol{\theta},s,a))$.

\theoremstyle{definition}
\begin{proposition}{} \textit{(Proof in the appendix)} Considering the expressions for $\sigma_{\text{epistemic}}$ and $\sigma_{\text{aleatoric}}$ in equations \ref{epistemicdef} and \ref{aleatoricdef},
\begin{align}
    \text{var}_{\boldsymbol{\theta},i }(y_i(\boldsymbol{\theta},s,a)) = \sigma_{\text{epistemic}}^2 + \sigma_{\text{aleatoric}}^2
\end{align}
\end{proposition}

We also consider two limit cases as sanity checks. First, in the absence of data, when all the uncertainty should be epistemic, we do find $\text{var}_{\boldsymbol{\theta},i}(y_i(\boldsymbol{\theta},s,a)) = \sigma_{\text{epistemic}}^2$. In the limit of infinite data, when all the uncertainty should be aleatoric, we also find $\text{var}_{\boldsymbol{\theta},i}(y_i(\boldsymbol{\theta},s,a)) = \sigma_{\text{aleatoric}}^2$.

\subsection{Approximate Uncertainties Using Two Networks}

\begin{figure}[h]
    \centering
    \includegraphics[width=0.45\textwidth]{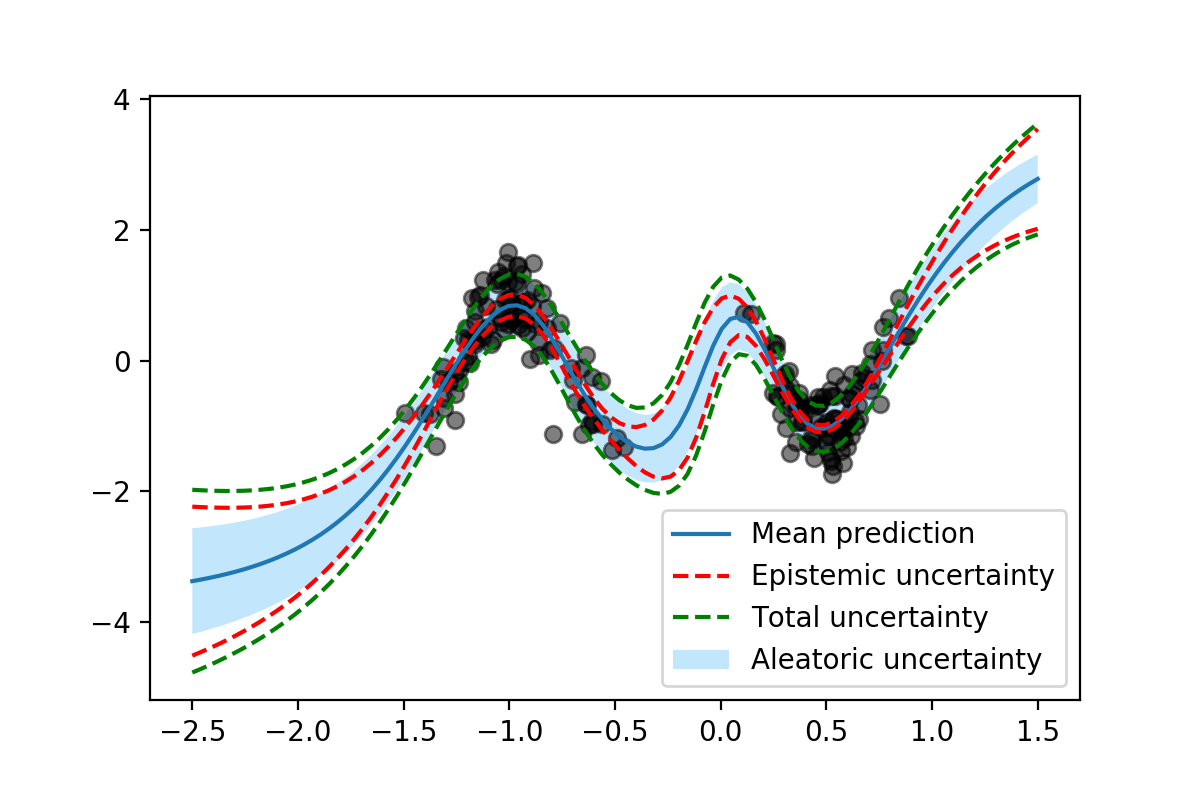}
    \vspace{-20pt}
    \caption{Illustration of uncertainty estimates provided by $\Tilde{\sigma}_\text{epistemic}$ and $\Tilde{\sigma}_\text{aleatoric}$ on a toy dataset (black dots). Intervals represent $\pm \sigma$ for all uncertainties, and the estimated total uncertainty $\Tilde{\sigma}_\text{total}$ is defined as $\Tilde{\sigma}_\text{total}^2 = \Tilde{\sigma}_\text{aleatoric}^2 + \Tilde{\sigma}_\text{epistemic}^2$.}
    \label{fig:illustration}
\end{figure}

Estimating the variance and expectation over $\boldsymbol{\theta}$ in the previous expressions for both uncertainties requires in principle a large number of samples of $\boldsymbol{\theta}$, which is impractical. Instead, we propose the following approximations of $\sigma_{\text{epistemic}}^2$ and $\sigma_{\text{aleatoric}}^2$ using only two samples $\boldsymbol{\theta}_A$ and $\boldsymbol{\theta}_B$ from the posterior distribution over $\boldsymbol{\theta}$,
\begin{align}
&\Tilde{\sigma}_\text{epistemic}^2 = \frac{1}{2} \E_{i \sim \mathcal{U}\{1,N\}} [ (y_i(\boldsymbol{\theta}_A,s,a) - y_i(\boldsymbol{\theta}_B,s,a))]^2 \nonumber  \\ 
&\Tilde{\sigma}_\text{aleatoric}^2 = \text{cov}_{i \sim \mathcal{U}\{1,N\}} ( y_i(\boldsymbol{\theta}_A,s,a),y_i(\boldsymbol{\theta}_B,s,a))
\label{uncertainty}
\end{align}

\theoremstyle{definition}
\begin{proposition}{}\textit{(Proof in the appendix)} $\Tilde{\sigma}_\text{epistemic}$ and $\Tilde{\sigma}_\text{aleatoric}$ are unbiased estimators of $\sigma_{\text{epistemic}}$ and $\sigma_{\text{aleatoric}}$. Moreover, assuming that the network outputs are uncorrelated, the variance of these estimators converges towards 0 as the number of quantiles increases.
\end{proposition}

In figure \ref{fig:illustration}, we provide an illustration of the uncertainties measured with $\Tilde{\sigma}_\text{epistemic}$ and $\Tilde{\sigma}_\text{aleatoric}$ on a toy dataset. We consider a neural network that estimates 50 quantiles from the target distribution, and we draw two samples of $\boldsymbol{\theta}$ using approximate MAP sampling \cite{pearce2018bayesian}. As expected, $\Tilde{\sigma}_\text{epistemic}$ is small close to the data but large far from it, while $\Tilde{\sigma}_\text{aleatoric}$ correctly captures the noise in the data.

\subsection{Uncertainty-Aware Deep Q Networks}

\setlength{\textfloatsep}{10pt}
\begin{algorithm}[ht]
\caption{UA-DQN action selection}
\begin{algorithmic}
\STATE {\bfseries Requires:} Action set $\mathcal{A}$, hyperparameters $\lambda$ and $\beta$, value network $\boldsymbol{\theta}_v$, and two auxiliary networks $\boldsymbol{\theta}_A$ and $\boldsymbol{\theta}_B$ approximately sampled from the posterior distribution over $\boldsymbol{\theta}$ with randomized MAP sampling \cite{pearce2018bayesian}.
\FOR{$a$ in $\mathcal{A}$}
\STATE Calculate action mean $\mu = \E_{i}[y_i(\boldsymbol{\theta}_v)]$
\STATE Calculate uncertainties $\Tilde{\sigma}_\text{epistemic}^2$ and $\Tilde{\sigma}_\text{aleatoric}^2$ using networks $\boldsymbol{\theta}_A$ and $\boldsymbol{\theta}_B$.
\STATE Adjust for risk-aversion: $\mu \leftarrow \mu - \lambda \Tilde{\sigma}_\text{aleatoric}$
\STATE Draw a sample $\hat{Q}_a$ from $\mathcal{N}(\mu,\beta \Tilde{\sigma}_\text{epistemic}^2)$
\ENDFOR
\STATE {\bfseries Output:} $\text{argmax}_a[\hat{Q}_a]$
\end{algorithmic}
\label{alg1}
\end{algorithm}

Until now, we have been mainly concerned with learning the return distribution given an ensemble of samples of this distribution. For temporal difference learning, we replace these samples with the Bellman target. Although this implies measuring the ``one-step" epistemic uncertainty on the bootstrapped target instead of that on the total return, this uncertainty is nonetheless useful for exploration as it allows for the identification of less-visited state-action pairs.

There are several ways these uncertainty estimates could be included into a reinforcement learning algorithm, for example to drive information-directed exploration \cite{nikolov2019information}. To better contrast the different roles played by both uncertainties, we propose a simple uncertainty-aware Deep Q Networks algorithm (UA-DQN), which is based on the QR-DQN algorithm of \cite{dabney2017distributional} but includes the following modifications, presented in Algorithm 1:

\textbf{Auxiliary networks for uncertainty estimation.} To disentangle value learning and uncertainty estimation, we consider two auxiliary networks $\boldsymbol{\theta}_A$ and $\boldsymbol{\theta}_B$ both trained on the targets used in QR-DQN and approximately sampled from the posterior distribution over $\boldsymbol{\theta}$. These networks are used to derive $\Tilde{\sigma}_\text{epistemic}$ and $\Tilde{\sigma}_\text{aleatoric}$. 

\textbf{Uncertainty-Aware Action Selection.} Instead of the $\epsilon$-greedy policy used by QR-DQN, we use our uncertainty estimates to separately drive risk-awareness and exploration. We use $\Tilde{\sigma}_\text{aleatoric}$ to penalize high-variance actions, while $\Tilde{\sigma}_\text{epistemic}$ drives exploration using Thompson sampling.

\section{Experiments}

\subsection{Safe Learning}

\begin{figure}[ht]
    \centering
    \includegraphics[width=0.3\textwidth]{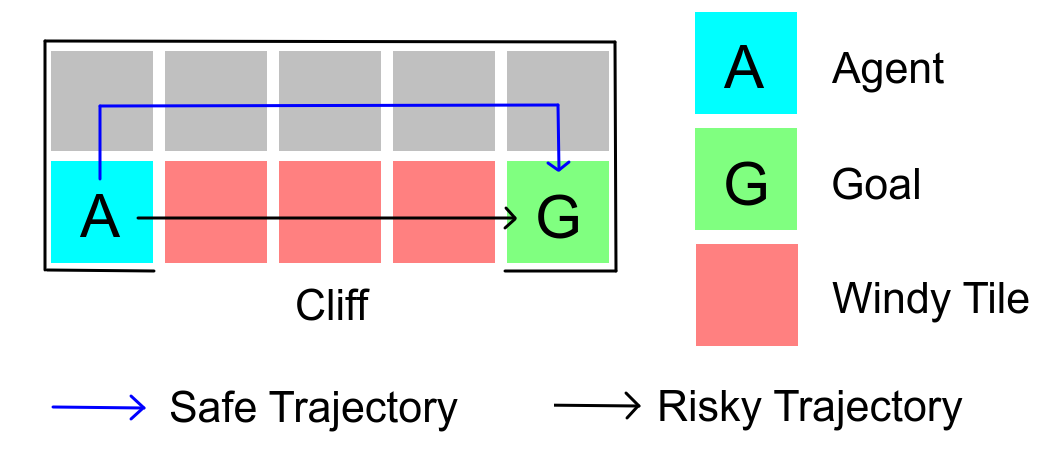}
    \includegraphics[width=0.45\textwidth]{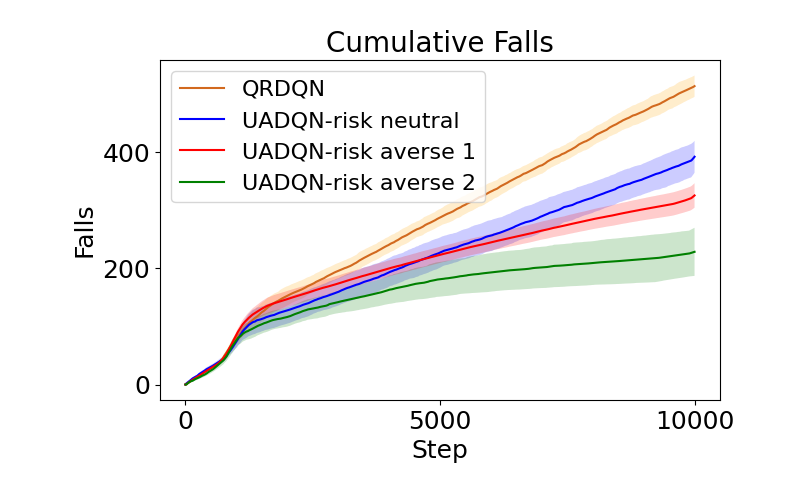}
    \vspace{-10pt}
    \caption{Top: grid environment. The risky trajectory has the highest expected reward but involves going through windy tiles where the agent may fall off the cliff. Bottom: Cumulative falls for different agents during training. Shaded areas indicate the 95\% confidence interval of the mean obtained from 30 training seeds.}
    \label{fig:gridworld}
\end{figure}
\begin{figure*}
    \centering
    \includegraphics[width=\textwidth]{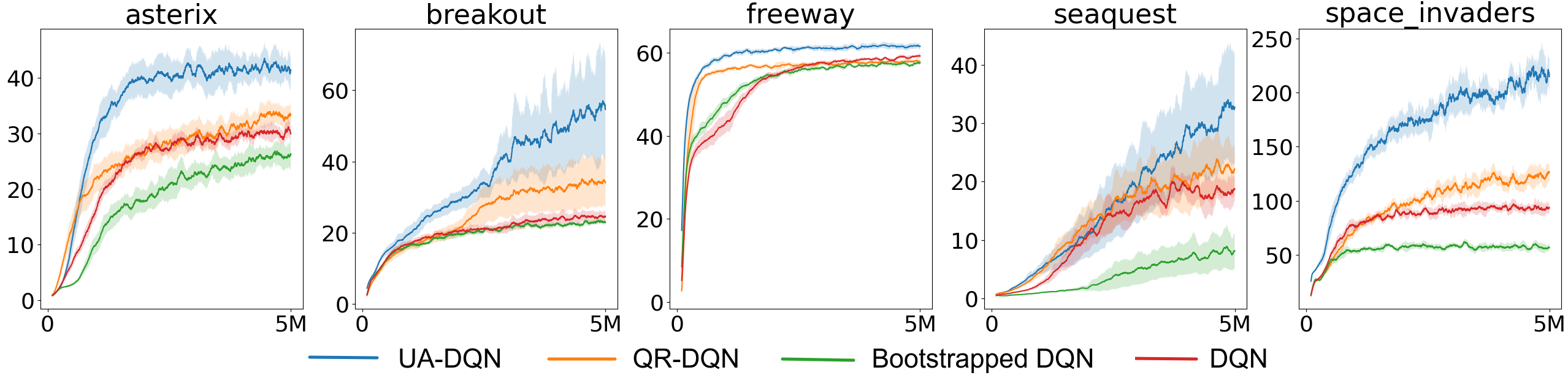}
    \vspace{-20pt}
    \caption{Learning curves over 5 million steps for different agents on the MinAtar testbed. Shaded areas correspond to the 95\% confidence interval of the mean obtained from 10 training seeds.}
    \label{fig:minatar}
\end{figure*}
We first empirically study the behavior of our uncertainty-aware DQN on an environment inspired by the AI Safety Gridworlds \cite{leike2017ai}. We consider a simple $2\times5$ gridworld represented in figure \ref{fig:gridworld}, in which the agent must navigate to a goal without falling off a cliff. The agent receives -1 point at each timestep, +10 points for reaching the goal, and if the agent falls off the cliff the environment restarts. We introduce a stochastic wind in this environment, which with probability $5\%$ knocks the agent off the cliff if the agent is on the ledge. The expectation value of the returns associated with the risky trajectory along the ledge is 4.8, while for the safe trajectory the return is deterministically 4. 

We compare the learning behavior of our uncertainty-aware DQN agent to other comparable algorithms. We consider the $\epsilon$-greedy QR-DQN algorithm of \cite{dabney2017distributional}, a risk-neutral version of UA-DQN (with $\lambda=0$), and two risk-averse variants of UA-DQN (with $\lambda=0.5$). Variant 1 uses the variance of the learned quantiles to estimate aleatoric uncertainty, while variant 2 uses our $\Tilde{\sigma}_\text{aleatoric}$ estimator. All algorithms differ only in action selection.

Experimental results are shown in figure \ref{fig:gridworld}. While all algorithms quickly learn to solve this simple task, there are marked differences in behavior. QR-DQN falls off the most due to both taking the risky trajectory and its $\epsilon$-greedy policy. Risk-neutral UA-DQN only falls off the cliff due to its use of the risky trajectory. Both risk-averse variants of UA-DQN learn to use the safe trajectory. However, variant 1 overestimates aleatoric uncertainty due to its use of a biased estimator, takes longer to identify the safe trajectory, and thus accumulates more falls during learning than variant 2 that uses our unbiased estimate $\Tilde{\sigma}_\text{aleatoric}$. Variant 2, which uses our unbiased estimators of both uncertainties, is the least likely to fall off the cliff during learning.

\subsection{Evaluation on MinAtar}

We now evaluate our UA-DQN algorithm on the MinAtar testbed \cite{young2019minatar}, which contains simplified implementations of 5 Atari games. Compared to the Arcade Learning Environment \cite{bellemare2013arcade}, MinAtar has similar underlying game dynamics but involves lower-dimensional observations. This helps to decouple representation learning from behavioral learning and allows to focus on the latter, and also encourages reproducibility as the reduced computational overhead allows for more thorough comparisons involving more training seeds.

We compare risk-neutral UA-DQN ($\lambda=0$) with DQN \cite{mnih2015human}, QR-DQN, and Bootstrapped DQN \cite{osband2016deep}. In contrast to UA-DQN which uses Thompson sampling to explore, Bootstrapped DQN uses an ensemble of bootstrapped DQN heads to achieve diverse behaviors.  All algorithms are implemented within the same code base using the hyperparameters of \cite{young2019minatar}, except that we use the Adam optimizer \cite{kingma2014adam} instead of RMSProp with learning rate $10^{-4}$ and $\epsilon=10^{-8}$. We also optimized the exploration hyperparameters: we use a final $\epsilon$ of 0.03 for the $\epsilon$-greedy policies of DQN, QR-DQN, and Bootstrapped DQN, and for UA-DQN we use $\beta=0.2$. We selected these values using the game Breakout, which were then fixed for all the other games.

Our results are shown in figure \ref{fig:minatar}. As reported in \cite{dabney2017distributional}, we find that QR-DQN outperforms the other DQN variants, and that UA-DQN in turn significantly outperforms QR-DQN. To understand why this is, we inspect the behavior of UA-DQN and find that even at the end of training roughly 10-20\% of the actions selected by UA-DQN are non-greedy. As UA-DQN only differs from QR-DQN in action selection, and QR-DQN's performance decreases with higher levels of $\epsilon$-greedy exploration, this result indicates that UA-DQN successfully uses  $\Tilde{\sigma}_\text{epistemic}$ to appropriately decide when best to perform exploratory or greedy actions. 

\section{Conclusion}

Estimating both uncertainties is important for developing agents that can both explore efficiently and account for risk in their actions. We propose a scheme whereby both types of uncertainty on the expected return of a policy can be estimated in deep reinforcement learning. We show that unbiased estimators for these uncertainties can be obtained using only two networks, and that these estimators can be efficiently harnessed by an uncertainty-aware DQN algorithm for improved risk-sensitivity and exploration. We find that this UA-DQN algorithm significantly outperforms other DQN variants on the MinAtar testbed.

\bibliography{references}
\bibliographystyle{icml2020}

\newpage
\onecolumn
\appendix

\section{Related Work}

Our work focuses on the problem of estimating the uncertainty of the expected return of a policy in model-free reinforcement learning. The epistemic uncertainty on the expected return has been shown to be useful for exploration \cite{osband2016deep,azizzadenesheli2018efficient,touati2018randomized}. On the other hand, the aleatoric uncertainty of the expected return is useful for designing risk-averse policies \cite{howard1972risk,tamar2016learning,dabney2018implicit}. Whereas most prior work considers both uncertainties separately, \cite{tang2018exploration,moerland2017efficient} are interested in both, but their methods yield only an aggregate uncertainty. \cite{nikolov2019information} do make use of both uncertainties to drive information-directed sampling \cite{russo2014learning}, but their uncertainty estimates derive from two different frameworks. Moreover, we argue that the variance of the learned return distribution, which \cite{nikolov2019information} use for aleatoric uncertainty estimation, conflates both uncertainties for out of distribution data. Our work aims to provide a single framework for simultaneously estimating both uncertainties for the return distribution.

Estimating both types of uncertainty is also important in model-based reinforcement learning, where uncertainties affect the predictions of a learned dynamics model of the environment. Uncertainty estimates can be used in planning, either for better exploration \cite{schmidhuber1991possibility, sun2011planning} or to avoid risky or unknown sections of the environment \cite{garcia2015comprehensive}. Model based algorithms that explicitly account for both aleatoric and epistemic uncertainties have recently also been developed \cite{depeweg2018decomposition,chua2018deep,henaff2019model}. 

An approach that combines model free and model based techniques consists of using the uncertainties derived from a learned dynamics model to inform the policy of a model-free agent. The epistemic uncertainty associated with the learned model can for example be used as an intrinsic motivation bonus \cite{stadie2015incentivizing,pathak2017curiosity,burda2018exploration}. However, uncertainties on the transition model do not typically convey information about the uncertainty of the expected return of a policy, which is a quantity of fundamental interest in reinforcement learning.

\section{Proofs}

In the following, for notational simplicity we will omit the dependence of $y_i(\boldsymbol{\theta},s,a)$ on $s$ and $a$. Moreover, subscripts used in variances/expectation values should be interpreted as the variance/expectation value taken over the distribution of the variables in the subscript, so that for example $\E_{\boldsymbol{\theta}} = \E_{\boldsymbol{\theta} \sim P(\boldsymbol{\theta}|D)}$ and $\E_i = \E_{i \sim \mathcal{U}\{1,N\}}$. We will also assume that the following integrals over $P(\boldsymbol{\theta}|D)$ are well defined, which, considering in particular the Gaussian prior over the weights, is a reasonable assumption.

\subsection{Proof of proposition 2.1}

Here, we show that, considering a sample $\hat{\boldsymbol{\theta}}$ drawn from the posterior distribution over $\boldsymbol{\theta}$, $\text{var}_{i}[y_i(\hat{\boldsymbol{\theta}})]$ is a biased estimator of $\sigma_{\text{aleatoric}}^2$. We do so by showing that $\E_{\boldsymbol{\theta}}[\text{var}_{i}[y_i(\boldsymbol{\theta})]]$ is greater than $\sigma_{\text{aleatoric}}^2$.
\begin{align*}
    \E_{\boldsymbol{\theta}} \left[ \text{var}_{i}[y_i(\boldsymbol{\theta},s,a)] \right] &= \E_{\boldsymbol{\theta}} \left[ \frac{1}{N} \sum_{j=1}^N (y_j(\boldsymbol{\theta})-\E_i[y_i(\boldsymbol{\theta})])^2 \right] \\
    &= \frac{1}{N} \sum_{j=1}^N \E_{\boldsymbol{\theta}} \left[ (y_j(\boldsymbol{\theta})-\E_i[y_i(\boldsymbol{\theta})])^2\right]
\end{align*}

By definition of the variance, we also have
\begin{align*}
    \text{var}_{\boldsymbol{\theta}}[y_j(\boldsymbol{\theta}) - \E_i (y_i(\boldsymbol{\theta})] = \E_{\boldsymbol{\theta}} \left[ (y_j(\boldsymbol{\theta})-\E_i[y_i(\boldsymbol{\theta})])^2\right] - (\E_{\boldsymbol{\theta}}[y_j(\boldsymbol{\theta})]-\E_{\boldsymbol{\theta},i}[y_i(\boldsymbol{\theta})])^2
\end{align*}

Therefore, when the posterior over $\boldsymbol{\theta}$ is not concentrated and $\text{var}_{\boldsymbol{\theta}}[y_j(\boldsymbol{\theta}) - \E_i (y_i(\boldsymbol{\theta})]>0$,
\begin{align*}
    \E_{\boldsymbol{\theta}} \left[ \text{var}_{i}[y_i(\boldsymbol{\theta},s,a)] \right] &> \frac{1}{N} \sum_{j=1}^N (\E_{\boldsymbol{\theta}}[y_j(\boldsymbol{\theta})]-\E_{\boldsymbol{\theta},i}[y_i(\boldsymbol{\theta})])^2 \\
    &> \text{var}_i [\E_{\boldsymbol{\theta}}[y_i(\boldsymbol{\theta})]] \\
    &> \sigma_{\text{aleatoric}}^2
\end{align*}

\subsection{Proof of proposition 2.2}

Here, we show that $\text{var}_{\boldsymbol{\theta},i}(y_i(\boldsymbol{\theta},s,a)) = \sigma_{\text{epistemic}}^2 + \sigma_{\text{aleatoric}}^2$. 

\begin{align*}
    \text{var}_{\boldsymbol{\theta},i}(y_i(\boldsymbol{\theta})) &= \int_{\boldsymbol{\theta}} \frac{1}{N} \sum_{j=1}^N \left(y_j(\boldsymbol{\theta}) - \E_{\boldsymbol{\theta},i} [y_i(\boldsymbol{\theta})]\right)^2 P(\boldsymbol{\theta}|D)  d\boldsymbol{\theta} \\
    &= \int_{\boldsymbol{\theta}} \frac{1}{N} \sum_{j=1}^N \left(y_j(\boldsymbol{\theta}) - \E_{\boldsymbol{\theta}} [y_j(\boldsymbol{\theta})] + \E_{\boldsymbol{\theta}} [y_j(\boldsymbol{\theta})] - \E_{\boldsymbol{\theta},i} [y_i(\boldsymbol{\theta})]\right)^2 P(\boldsymbol{\theta}|D) d\boldsymbol{\theta} \\
    &= \int_{\boldsymbol{\theta}} \frac{1}{N} \sum_{j=1}^N \left( \left(y_j(\boldsymbol{\theta}) - \E_{\boldsymbol{\theta}} [y_j(\boldsymbol{\theta})]\right)^2 \right.
    \\& \qquad + ( \E_{\boldsymbol{\theta}} [y_j(\boldsymbol{\theta})] - \E_{\boldsymbol{\theta},i} [y_i(\boldsymbol{\theta})])^2 
    \\& \qquad + 2 ( \E_{\boldsymbol{\theta}} [y_j(\boldsymbol{\theta})] - \E_{\boldsymbol{\theta},i} [y_i(\boldsymbol{\theta})])(y_j(\boldsymbol{\theta}) - \E_{\boldsymbol{\theta}} [y_j(\boldsymbol{\theta})]) \Big ) P(\boldsymbol{\theta}|D) d\boldsymbol{\theta}
\end{align*}
The integral over $\boldsymbol{\theta}$ of the last line is 0, which leaves us with
\begin{align*}
    \text{var}_{\boldsymbol{\theta},i}(y_i(\boldsymbol{\theta})) &= \int_{\boldsymbol{\theta}} \frac{1}{N} \sum_{j=1}^N \left(y_j(\boldsymbol{\theta}) - \E_{\boldsymbol{\theta}} [y_j(\boldsymbol{\theta})]\right)^2 P(\boldsymbol{\theta}|D) d\boldsymbol{\theta} + \int_{\boldsymbol{\theta}} \frac{1}{N} \sum_{j=1}^N ( \E_{\boldsymbol{\theta}} [y_j(\boldsymbol{\theta})] - \E_{\boldsymbol{\theta},i} [y_i(\boldsymbol{\theta})])^2 P(\boldsymbol{\theta}|D) d\boldsymbol{\theta}
    \\&=  \frac{1}{N} \sum_{j=1}^N \int_{\boldsymbol{\theta}} \left(y_j(\boldsymbol{\theta}) - \E_{\boldsymbol{\theta}} [y_j(\boldsymbol{\theta})]\right)^2 P(\boldsymbol{\theta}|D) d\boldsymbol{\theta} + \frac{1}{N} \sum_{j=1}^N ( \E_{\boldsymbol{\theta}} [y_j(\boldsymbol{\theta})] - \E_{\boldsymbol{\theta},i} [y_i(\boldsymbol{\theta})])^2
    \\&= \E_i(\text{var}_{\boldsymbol{\theta}}(y_i(\boldsymbol{\theta}))) + \text{var}_i ( \E_{\boldsymbol{\theta}} y_i(\boldsymbol{\theta}))
    \\&= \sigma_{\text{epistemic}}^2 + \sigma_{\text{aleatoric}}^2
\end{align*}

\subsection{Proof of proposition 2.3}

\subsection{Expectation of the estimators}

Here, we show that $\Tilde{\sigma}_\text{epistemic}$ and $\Tilde{\sigma}_\text{aleatoric}$ are unbiased estimators of $\sigma_{\text{epistemic}}$ and $\sigma_{\text{aleatoric}}$. In the following, $\E_{\boldsymbol{\theta_A},\boldsymbol{\theta_B}}$ indicates the expectation value when $\boldsymbol{\theta_A}$ and $\boldsymbol{\theta_B}$ are drawn from the posterior distribution over $\boldsymbol{\theta}$. Moreover, in what follows it can easily be verified that expectations over $\boldsymbol{\theta}$ and over $i$ are interchangeable due to the discrete nature of the expectation over $i$.

\begin{align*}
    \E_{\boldsymbol{\theta_A},\boldsymbol{\theta_B}} [\Tilde{\sigma}_\text{epistemic}^2] &= \frac{1}{2} \E_{\boldsymbol{\theta_A},\boldsymbol{\theta_B}}  \E_i [ (y_i(\boldsymbol{\theta}_A) - y_i(\boldsymbol{\theta}_B))^2] \\
    &= \frac{1}{2} \E_{\boldsymbol{\theta_A},\boldsymbol{\theta_B}}  \E_i [(y_i(\boldsymbol{\theta}_A) - \E_{\boldsymbol{\theta}}(y_i(\boldsymbol{\theta})) + \E_{\boldsymbol{\theta}}(y_i(\boldsymbol{\theta})) - y_i(\boldsymbol{\theta}_B))^2] \\
    &= \frac{1}{2} \E_{\boldsymbol{\theta_A},\boldsymbol{\theta_B}} \left[ \E_i [(y_i(\boldsymbol{\theta}_A) - \E_{\boldsymbol{\theta}}(y_i(\boldsymbol{\theta})))^2] +\E_i[ (\E_{\boldsymbol{\theta}}(y_i(\boldsymbol{\theta})) - y_i(\boldsymbol{\theta}_B))^2] \right.
    \\&\qquad \left. + 2 \E_i[( (\E_{\boldsymbol{\theta}}(y_i(\boldsymbol{\theta})) - y_i(\boldsymbol{\theta}_B))(y_i(\boldsymbol{\theta}_A) - \E_{\boldsymbol{\theta}}(y_i(\boldsymbol{\theta})))] \right]
\end{align*}
The average over either $\boldsymbol{\theta_A}$ or $\boldsymbol{\theta_B}$ of the last line is zero, which, after noticing that $\boldsymbol{\theta_A}$ and $\boldsymbol{\theta_B}$ are now separable such that we can use the equality $\E_{\boldsymbol{\theta_A}}[y_i(\boldsymbol{\theta}_A)] = \E_{\boldsymbol{\theta_B}}[y_i(\boldsymbol{\theta}_B)] = \E_{\boldsymbol{\theta}}[y_i(\boldsymbol{\theta})]$, leaves us with
\begin{align*}
    \E_{\boldsymbol{\theta_A},\boldsymbol{\theta_B}} [\Tilde{\sigma}_\text{epistemic}^2] &= \frac{1}{2}\left( \E_{\boldsymbol{\theta}} \left[ \E_i (y_i(\boldsymbol{\theta}) - \E_{\boldsymbol{\theta}}(y_i(\boldsymbol{\theta})))^2 + \E_i (\E_{\boldsymbol{\theta}}(y_i(\boldsymbol{\theta})) - y_i(\boldsymbol{\theta}))^2 \right] \right) \\
    &= \E_{\boldsymbol{\theta}} \left[ \E_i (y_i(\boldsymbol{\theta}) - \E_{\boldsymbol{\theta}}(y_i(\boldsymbol{\theta})))^2\right] \\
    &= \E_i \left[ \E_{\boldsymbol{\theta}} (y_i(\boldsymbol{\theta}) - \E_{\boldsymbol{\theta}}(y_i(\boldsymbol{\theta})))^2\right] \\
    &= \E_i \left[ \text{var}_{\boldsymbol{\theta}}(y_i(\boldsymbol{\theta})) \right] \\
    &= \sigma_\text{epistemic}^2
\end{align*}
so $\Tilde{\sigma}_\text{epistemic}$ is indeed an unbiased estimator of $\sigma_\text{epistemic}$.

Similarly, for $\Tilde{\sigma}_\text{aleatoric}$, and introducing $\epsilon_i(\boldsymbol{\theta}_A) = y_i(\boldsymbol{\theta}_A)-\E_{\boldsymbol{\theta}}(y_i(\boldsymbol{\theta}))$ and $\epsilon_i(\boldsymbol{\theta}_B) =  y_i(\boldsymbol{\theta}_B)-\E_{\boldsymbol{\theta}}(y_i(\boldsymbol{\theta}))$,

\begin{align*}
    \E_{\boldsymbol{\theta_A},\boldsymbol{\theta_B}} [\Tilde{\sigma}_\text{aleatoric}^2] &= \E_{\boldsymbol{\theta_A},\boldsymbol{\theta_B}} \text{cov}_i ( y_i(\boldsymbol{\theta}_A),y_i(\boldsymbol{\theta}_B)) \\
    &= \E_{\boldsymbol{\theta_A},\boldsymbol{\theta_B}} \text{cov}_i ( \epsilon_i(\boldsymbol{\theta}_A) + \E_{\boldsymbol{\theta}}(y_i(\boldsymbol{\theta})) ,\epsilon_i(\boldsymbol{\theta}_B) + \E_{\boldsymbol{\theta}}(y_i(\boldsymbol{\theta}))) \\
    &= \E_{\boldsymbol{\theta_A},\boldsymbol{\theta_B}} \left[\text{cov}_i(\E_{\boldsymbol{\theta}}(y_i(\boldsymbol{\theta})),\E_{\boldsymbol{\theta}}(y_i(\boldsymbol{\theta})))   + \text{cov}_i(\epsilon_i(\boldsymbol{\theta}_A),\E_{\boldsymbol{\theta}}(y_i(\boldsymbol{\theta}))) \right. \\
    &\qquad \left. + \text{cov}_i(\E_{\boldsymbol{\theta}}(y_i(\boldsymbol{\theta})),\epsilon_i(\boldsymbol{\theta}_B)) + \text{cov}_i(\epsilon_i(\boldsymbol{\theta}_A),\epsilon_i(\boldsymbol{\theta}_B)) \right] 
\end{align*}
Looking at these terms individually, we have
\begin{flalign*}
    \E_{\boldsymbol{\theta_A},\boldsymbol{\theta_B}} \left[\text{cov}_i(\E_{\boldsymbol{\theta}}(y_i(\boldsymbol{\theta})),\E_{\boldsymbol{\theta}}(y_i(\boldsymbol{\theta}))) \right] &= \text{var}_i(\E_{\boldsymbol{\theta}}(y_i(\boldsymbol{\theta}))) &\\
    &= \sigma_\text{aleatoric}^2
\end{flalign*}
\begin{flalign*}
    E_{\boldsymbol{\theta_A},\boldsymbol{\theta_B}} \left[\text{cov}_i(\epsilon_i(\boldsymbol{\theta}_A),\E_{\boldsymbol{\theta}}(y_i(\boldsymbol{\theta}))) \right] &= E_{\boldsymbol{\theta_A}} \left[\text{cov}_i(\epsilon_i(\boldsymbol{\theta}_A),\E_{\boldsymbol{\theta}}(y_i(\boldsymbol{\theta}))) \right] &\\
    &= E_{\boldsymbol{\theta_A}}\left[ \frac{1}{N} \sum_{j=1}^N (\epsilon_j(\boldsymbol{\theta}_A) - \E_i(\epsilon_i(\boldsymbol{\theta}_A)))(\E_{\boldsymbol{\theta}}(y_j(\boldsymbol{\theta})) - \E_i \E_{\boldsymbol{\theta}}(y_i(\boldsymbol{\theta}))) \right] \\
    &=  \frac{1}{N} \sum_{j=1}^N (\E_{\boldsymbol{\theta}}(y_j(\boldsymbol{\theta})) - \E_i \E_{\boldsymbol{\theta}}(y_i(\boldsymbol{\theta}))) ( \E_{\boldsymbol{\theta_A}} (\epsilon_j(\boldsymbol{\theta}_A)) - \E_i(\E_{\boldsymbol{\theta_A}}(\epsilon_i(\boldsymbol{\theta}_A))) )  \\
    &= 0 \qquad \text{since} \quad \E_{\boldsymbol{\theta_A}} (\epsilon_i(\boldsymbol{\theta_A})) = 0 \quad \text{for all } i
\end{flalign*}
\begin{flalign*}
    E_{\boldsymbol{\theta_A},\boldsymbol{\theta_B}} \left[\text{cov}_i(\E_{\boldsymbol{\theta}}(y_i(\boldsymbol{\theta}))),\epsilon_i(\boldsymbol{\theta}_B) \right] &= 0 \quad \text{[Same derivation as previous expression]} &
\end{flalign*}
\begin{flalign*}
    E_{\boldsymbol{\theta_A},\boldsymbol{\theta_B}} \left[\text{cov}_i(\epsilon_i(\boldsymbol{\theta}_A),\epsilon_i(\boldsymbol{\theta}_B)) \right] &=  E_{\boldsymbol{\theta_A},\boldsymbol{\theta_B}} \left[\frac{1}{N} \sum_{j=1}^N (\epsilon_j(\boldsymbol{\theta}_A) - \E_i(\epsilon_i(\boldsymbol{\theta}_A)))(\epsilon_j(\boldsymbol{\theta}_B) - \E_i(\epsilon_i(\boldsymbol{\theta}_B))) \right] &\\
    &= E_{\boldsymbol{\theta_A}}\left[\frac{1}{N} \sum_{j=1}^N (\epsilon_j(\boldsymbol{\theta}_A) - \E_i(\epsilon_i(\boldsymbol{\theta}_A)))(E_{\boldsymbol{\theta_B}}(\epsilon_j(\boldsymbol{\theta}_B)) - E_i(\E_{\boldsymbol{\theta_B}}(\epsilon_i(\boldsymbol{\theta}_B)))) \right] \\
    &= 0
\end{flalign*}
As desired, we end up with
\begin{align*}
    \E_{\boldsymbol{\theta_A},\boldsymbol{\theta_B}} [\Tilde{\sigma}_\text{aleatoric}^2] &= \sigma_\text{aleatoric}^2
\end{align*}

\subsection{Variance of the estimators}

Using the same notation as in the previous section, we can write
\begin{align*}
    \text{var}_{\boldsymbol{\theta_A},\boldsymbol{\theta_B}}[\Tilde{\sigma}_\text{epistemic}^2] &= \frac{1}{4} \text{var}_{\boldsymbol{\theta_A},\boldsymbol{\theta_B}}  \E_i [(y_i(\boldsymbol{\theta}_A) - y_i(\boldsymbol{\theta}_B))^2] \\
    &= \frac{1}{4} \text{var}_{\boldsymbol{\theta_A},\boldsymbol{\theta_B}} \left[ \frac{1}{N} \sum_{i=1}^N (y_i(\boldsymbol{\theta}_A) - y_i(\boldsymbol{\theta}_B))^2 \right] \\
    &= \frac{1}{4N^2}\text{var}_{\boldsymbol{\theta_A},\boldsymbol{\theta_B}} \left[ \sum_{i=1}^N (y_i(\boldsymbol{\theta}_A) - y_i(\boldsymbol{\theta}_B))^2 \right]
\end{align*}
We now require our assumption that all outputs of the neural networks are decorrelated to write
\begin{align*}
    \text{var}_{\boldsymbol{\theta_A},\boldsymbol{\theta_B}}[\Tilde{\sigma}_\text{epistemic}^2] &= \frac{1}{4N^2} \sum_{i=1}^N \text{var}_{\boldsymbol{\theta_A},\boldsymbol{\theta_B}} \left[  (y_i(\boldsymbol{\theta}_A) - y_i(\boldsymbol{\theta}_B))^2 \right] \\
    &= \frac{1}{4N^2} \sum_{i=1}^N \text{var}_{\boldsymbol{\theta_A},\boldsymbol{\theta_B}} \left[  y_i(\boldsymbol{\theta}_A)^2 + y_i(\boldsymbol{\theta}_B)^2 + 2y_i(\boldsymbol{\theta}_A)y_i(\boldsymbol{\theta}_B) \right] \\
    &\leq \frac{3}{4N^2} \sum_{i=1}^N  \left[ \text{var}_{\boldsymbol{\theta_A},\boldsymbol{\theta_B}} [  y_i(\boldsymbol{\theta}_A)^2] + \text{var}_{\boldsymbol{\theta_A},\boldsymbol{\theta_B}}[y_i(\boldsymbol{\theta}_B)^2] + 4\text{var}_{\boldsymbol{\theta_A},\boldsymbol{\theta_B}}[y_i(\boldsymbol{\theta}_A)y_i(\boldsymbol{\theta}_B) ] \right] \\
    & \qquad \text{[Where we used the Cauchy-Schwartz inequality]} \\
    &\leq \frac{3}{4N^2} \sum_{i=1}^N \left[ 2 \text{var}_{\boldsymbol{\theta}} [  y_i(\boldsymbol{\theta})^2] + 8(\E_{\boldsymbol{\theta}}y_i(\boldsymbol{\theta}))^2 \text{var}_{\boldsymbol{\theta}}[y_i(\boldsymbol{\theta})] +  2(\text{var}_{\boldsymbol{\theta}}[y_i(\boldsymbol{\theta})])^2 \right]
\end{align*}
We now further assume that $\E_{\boldsymbol{\theta}}[y_i(\boldsymbol{\theta})]$, $\text{var}_{\boldsymbol{\theta}}[y_i(\boldsymbol{\theta})]$, and $\text{var}_{\boldsymbol{\theta}}[y_i^2(\boldsymbol{\theta})]$ are bounded for all $i$ and $N$. Then, there is a constant $C$ such that, for all $i$ and $N$,
\begin{align*}
2 \text{var}_{\boldsymbol{\theta}} [  y_i(\boldsymbol{\theta})^2] + 8(\E_{\boldsymbol{\theta}}[y_i(\boldsymbol{\theta})])^2 \text{var}_{\boldsymbol{\theta}}[y_i(\boldsymbol{\theta})] +  2(\text{var}_{\boldsymbol{\theta}}[y_i(\boldsymbol{\theta})])^2 \leq C
\end{align*}
We then obtain
\begin{align*}
    \text{var}_{\boldsymbol{\theta_A},\boldsymbol{\theta_B}}[\Tilde{\sigma}_\text{epistemic}^2] &\leq  \frac{3}{4N^2} \sum_{i=1}^N C \\
    &\leq \frac{C}{4N}
\end{align*}
The variance of $\Tilde{\sigma}_\text{epistemic}^2$ (and thus that of $\Tilde{\sigma}_\text{epistemic}$) therefore decreases towards 0 as the number of quantiles increases.

As for the aleatoric uncertainty, a similar bound can be derived by rewriting $\text{var}_{\boldsymbol{\theta_A},\boldsymbol{\theta_B}}[\Tilde{\sigma}_\text{aleatoric}^2]$ as follows.
\begin{align*}
    \text{var}_{\boldsymbol{\theta_A},\boldsymbol{\theta_B}}[\Tilde{\sigma}_\text{aleatoric}^2] &= \text{var}_{\boldsymbol{\theta_A},\boldsymbol{\theta_B}} [\text{cov}_i ( y_i(\boldsymbol{\theta}_A),y_i(\boldsymbol{\theta}_B))] \\
    &= \text{var}_{\boldsymbol{\theta_A},\boldsymbol{\theta_B}} \left[ \frac{1}{N} \sum_{j=1}^N (y_j(\boldsymbol{\theta}_A) - \E_i y_i(\boldsymbol{\theta}_A))(y_j(\boldsymbol{\theta}_B) - \E_i y_i(\boldsymbol{\theta}_B)) \right] \\
    &= \frac{1}{N^2} \text{var}_{\boldsymbol{\theta_A},\boldsymbol{\theta_B}} \left[  \sum_{j=1}^N (y_j(\boldsymbol{\theta}_A) - \E_i y_i(\boldsymbol{\theta}_A))(y_j(\boldsymbol{\theta}_B) - \E_i y_i(\boldsymbol{\theta}_B)) \right]
\end{align*}
In a manner similar as for the derivation of the variance of $\Tilde{\sigma}_\text{epistemic}^2$, assuming that the network outputs are uncorrelated and that the first moments of $y_i(\boldsymbol{\theta})$ are bounded, we can also derive a bound for $\text{var}_{\boldsymbol{\theta_A},\boldsymbol{\theta_B}}[\Tilde{\sigma}_\text{aleatoric}^2]$ that converges to 0 with increasing $N$.

\section{Correlations between the outputs of a Bayesian neural network}

Proposition 2.3 makes the assumption that the network outputs are uncorrelated. Indeed, correlations between outputs could cause for example a network to overestimate \textit{all} the quantiles. If both networks A and B produce overestimations, then $\Tilde{\sigma}_\text{epistemic}$ would probably underestimate $\sigma_{\text{epistemic}}$. However, in the limit of infinite width Bayesian neural networks are uncorrelated for normal priors and separable likelihoods \cite{neal2012bayesian}. In the following, we experimentally explore in which cases this applies to finite width neural networks and to approximate Bayesian techniques such as the randomized MAP sampling technique \cite{pearce2018bayesian} used in our work.

\subsection{Uncertainties for different network widths}

\begin{figure*}[ht]
    \centering
    \includegraphics[width=0.44\textwidth]{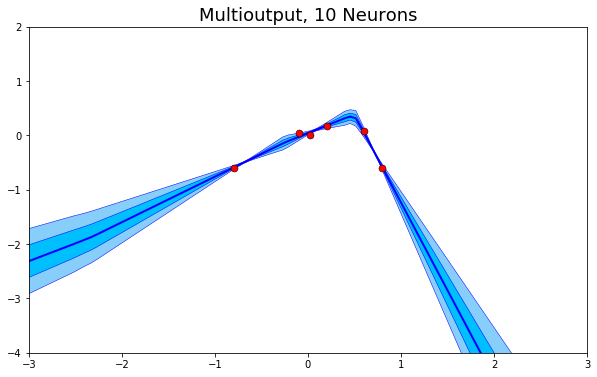}
    \includegraphics[width=0.44\textwidth]{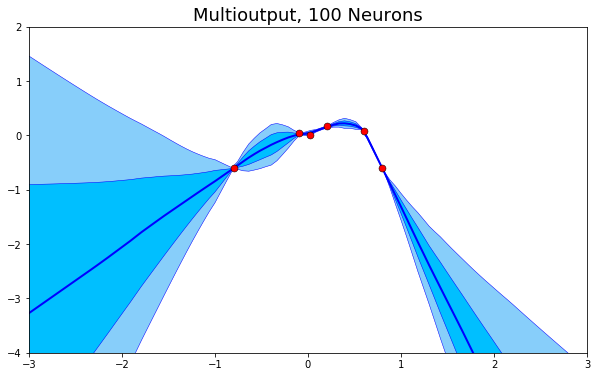}
    \includegraphics[width=0.44\textwidth]{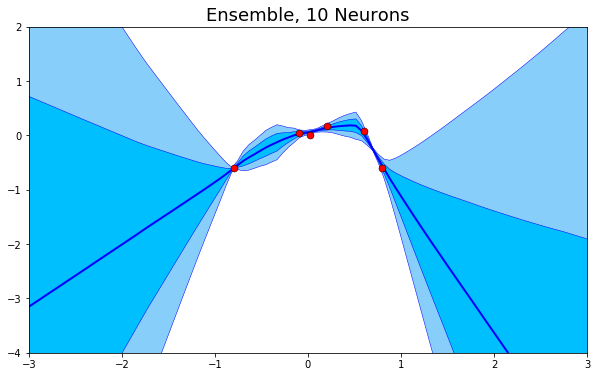}
    \includegraphics[width=0.44\textwidth]{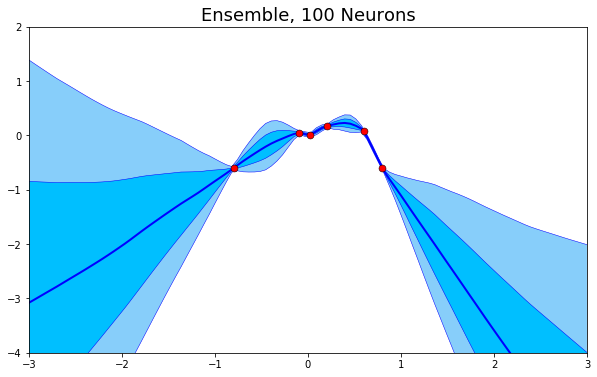}
    \caption{Comparison of epistemic uncertainties obtained with the approximate MAP sampling method of \cite{pearce2018bayesian}. Top: uncertainties obtained by a single neural network with twenty outputs. Bottom: uncertainties obtained by an ensemble of 20 networks. Left: 10 neurons per hidden layer. Right: 100 neurons per hidden layer. The two colors of shading represent one and two standard deviations from the mean.}
    \label{fig:multioutput}
    \vspace{10pt}
\end{figure*}

First, we compare the epistemic uncertainties produced by an ensemble of neural networks produced by the "anchoring" approximate MAP sampling technique of \cite{pearce2018bayesian} to that produced by a single neural network (also produced with approximate MAP sampling) with several outputs on a toy regression problem. Both the problem formulation and the code for this experiment draw from the work of \cite{pearce2018bayesian}.

Representative samples from these experiments are shown in figure \ref{fig:multioutput}. For a small neural network with only 10 neurons per layer the different outputs of the multioutput neural network are indeed strongly correlated, which leads to poor uncertainty estimates (top left). The ensemble produces significantly better uncertainty estimates for the same network width (bottom left). However, as we increase the width of the neural network to 100 (top right) the uncertainty estimates of the network with multiple outputs improve and become close to those obtained by the larger ensemble of networks of the same width (bottom right).

\section{Further information on the MinAtar experiment}

Our MinAtar experiments used the same network structure as that used in \cite{young2019minatar} and, apart from the optimized exploration hyperparameters and our use of the Adam optimizer described in the main text, also the same hyperparameters indicated in table \ref{tab:minatar hyperparams}. We searched among $\{10^{-4},2.5\times 10^{-4}\}$ for the Adam learning rate and $\{10^{-8},0.01/32\}$ for Adam $\epsilon$, and among $\{0.1,0.03,0.01\}$ for final exploration $\epsilon$ using QR-DQN on Breakout. We found that whereas 0.01 and 0.03 lead to similar average scores, a value of 0.03 led to smaller variance in the results. For UA-DQN, we searched among $\{0.5,0.2,0.1\}$ for $\beta$ on Breakout.

To approximately sample from the posterior over $\boldsymbol{\theta}$ for the auxiliary networks used in UA-DQN, we use the approximate MAP sampling scheme of \cite{pearce2018bayesian}. For this scheme, we set the scale of the noise to a realistic value of 1, and the scale of the prior to the standard deviation of the network weights at initialization.

\begin{table}[ht]
\vspace{20pt}
    \centering
    \begin{tabular}{l|l}
        Hyperparameter & Value \\ \hline
        minibatch size & 32 \\
        replay buffer size & 100000 \\
        target network update frequency & 1000 \\
        discount factor & 0.99 \\
        number of step & 5000000 \\
        Adam learning rate & $10^{-4}$ \\
        Adam $\epsilon$ & $10^{-8}$ \\
        replay start size & 5000 \\
        update frequency & 1 \\
        initial $\epsilon$ \textit{(DQN, QR-DQN, Bootstrapped DQN)} & 1 \\ 
        final $\epsilon$ \textit{(DQN, QR-DQN, Bootstrapped DQN)} & 0.03 \\
        final exploration step \textit{(DQN, QR-DQN, Bootstrapped DQN)} & 100000 \\
        Bootstrapped heads \textit{(Bootstrapped DQN)} & 10 \\
        Number of quantiles \textit{(QR-DQN, UA-DQN)} & 50 \\
        $\beta$ \textit{(UA-QDN)} & 0.2 \\
        $\lambda$ \textit{(UA-QDN)} & 0
    \end{tabular}
    \label{tab:minatar hyperparams}
    \caption{Hyperparameters used for our MinAtar experiments}
\end{table}

\end{document}